\pdfoutput=1
\documentclass[compsoc,journal]{IEEEtran}

\usepackage{latexsym}
\usepackage{graphicx}
\usepackage{amsthm}  
\usepackage{array}
\usepackage{multirow}
\newcolumntype{C}[1]{>{\centering\let\newline\\\arraybackslash\hspace{0pt}}m{#1}}
\usepackage{tikz}
\def\checkmark{\tikz\fill[scale=0.45](0,.35) -- (.25,0) -- (0.8,.7) -- (.25,.15) -- cycle;} 

\begin{document}
\newcommand{\citep}{\cite}
\newcommand{\citet}{\cite}

\newcommand{\revcom}{}

\newtheorem{theorem}{Theorem}
\newtheorem{thFS}{Theorem}
\newtheorem{thProof}{Theorem}

\title{Automatic Synthesis of \\ Totally Self-Checking Circuits}

\author{Michael~Garvie,~\IEEEmembership{Member,~IEEE,} Phil~Husbands\thanks{The authors are with the Department of Informatics, University of Sussex, Falmer, BN1 9RH, UK.  Email: M.Garvie@sussex.ac.uk.  This research was made possible by processor time donated by the distributed computing community and some experiments presented in this paper were carried out using the Grid'5000 testbed, supported by a scientific interest group hosted by Inria and including CNRS, RENATER and several Universities as well as other organizations.}}

\maketitle

\begin{abstract}
Totally self-checking (TSC) circuits are synthesised with a grid of computers running a distributed population based stochastic optimisation algorithm. The presented method is the first to automatically synthesise TSC circuits from arbitrary logic as all previous methods fail to guarantee the checker is self-testing (ST) for circuits with limited output codespaces. The circuits synthesised by the presented method have significantly lower overhead than the previously reported best for every one of a set of 11 frequently used benchmarks.  Average overhead across the entire set is 23\% of duplication and comparison overhead, compared with an average of 69\% for the previous best reported values across the set. The methodology presented represents a breakthrough in concurrent error detection (CED). The highly efficient, novel designs produced are tailored to each circuit's function, rather than being constrained by a particular modular CED design methodology. Results are synthesised using two-input gates and are TSC with respect to all gate input and output stuck-at faults. The method can be used to add CED with or without modifications to the original logic, and can be generalised to any implementation technology and fault model. An example circuit is analysed and rigorously proven to be TSC.
\end{abstract}

\IEEEpeerreviewmaketitle


\section{Introduction}
\label{digSC}
\label{intro}
\begin{figure}
\centering
\includegraphics[width=6.5cm]{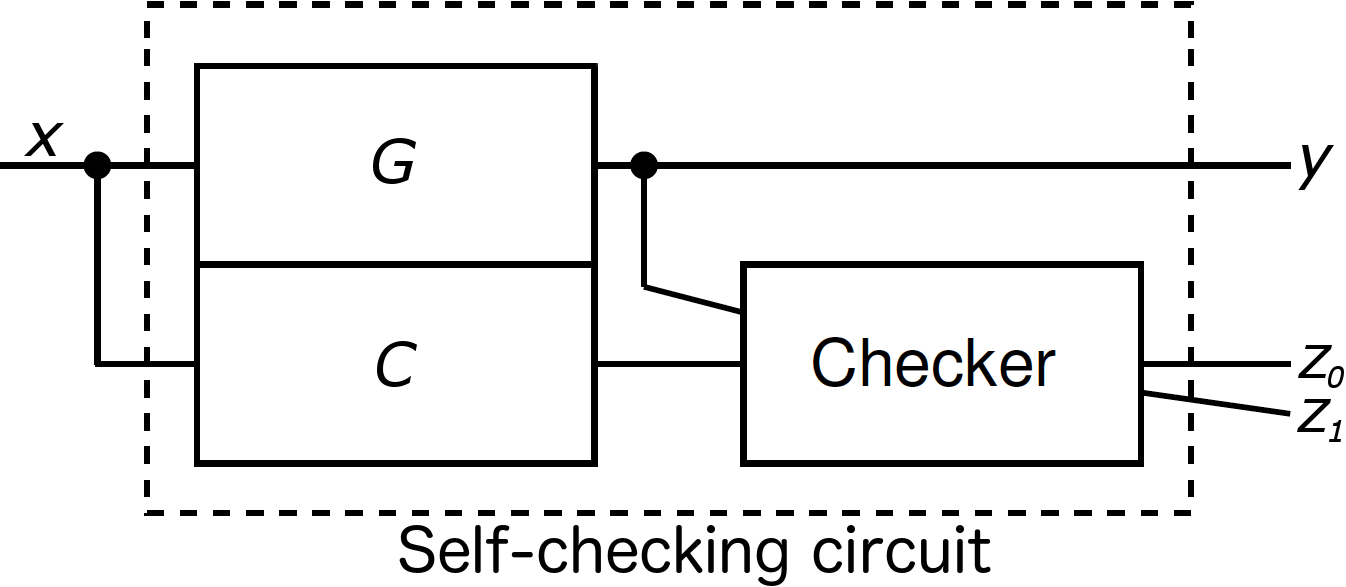}
\caption{Most self-checking circuit synthesis techniques generate three modules: one to generate primary output function $G$, another generating check bits $C(x)$ and a third to check whether $G(x)C(x)$ forms a valid codeword.  The overall self-checking circuit outputs a dual rail error signal $z$ and POs $y$.  Until now it has been an open question whether non-modular TSC circuits  -- ie. with all barriers within the dotted lines  blurred out -- exist for arbitrary functions and whether an automatic synthesis process could be built to find them.}
\label{fig2mod}
\end{figure}

\IEEEPARstart{T}{otally} self-checking (TSC) circuits \cite{cart:68,ande:71} (defined in \S\ref{defnot}) guarantee correct output until they signal an error, under certain fault assumptions.  Some self-checking circuit synthesis techniques (Table \ref{tablecompetition}) sometimes produce TSC circuits \cite{toub:97,das:99,ghos:05,mitr:00,de:94} while others offer partial fault coverage \cite{dutt:05,efan:16,goes:00,gess:03,moro:00}.  Some add concurrent error detection (CED) without requiring modifications to the original circuit (ie. are non-intrusive) \cite{leve:90,pare:95,dutt:05,dali:13,goes:00,gess:03,moro:00} whilst others impose structural restraints on the primary output (PO) function generating logic \cite{stan:03,toub:97,das:99,ghos:05}.  Most techniques encode outputs using extra bits (without altering PO encoding in the case of systematic codes) and require a checker to signal an error when the code is broken (Fig. \ref{fig2mod}).  Table \ref{tablecompetition} summaries the properties of state-of-the-art self-checking circuit synthesis techniques in terms of the attributes mentioned above, and includes the two variants (unconstrained and non-intrusive) of the novel method introduced in this paper.

\begin{table*}
\centering
\begin{tabular}{p{1.15cm} || p{0.75cm} p{0.75cm} p{0.55cm} p{0.75cm} p{0.75cm} p{0.75cm} p{0.55cm} p{0.55cm} p{0.55cm} p{0.65cm} p{0.80cm} | p{0.75cm} p{0.75cm}}
\centering
	& \cite{stan:03}	& \cite{toub:97}	& \cite{das:99}	& \cite{ghos:05}		& \cite{dutt:05}	& \cite{efan:16}& \cite{dali:13}	& \cite{goes:00}	& \cite{gess:03}	& \cite{moro:00}	& \cite{mitr:00} 	& $E_U$ 	& $E_{NI}$ \\
\hline
\hline
Method / \newline Encoding & Berger/ \newline Bose-Lin/ \newline Parity & Parity groups & Bose-Lin & Parity groups &Even parity& Weight based & Parity groups & 1-out-of-4 & 1-out-of-3 & Berger & Parity/ Berger/ Bose-Lin & Evolved ad hoc & Evolved ad hoc \\
\hline
H\&I &			& \checkmark		& 			& \checkmark			&			&			& \checkmark		&			&			&				&			& \checkmark 	& \checkmark \\
\hline
NI & 	&			&			&				& \checkmark		&			& \checkmark		& \checkmark		& \checkmark		& \checkmark			&			&		& \checkmark \\
\hline
FS	&	\checkmark		& \checkmark		& \checkmark		& \checkmark			&			&			& \checkmark		& 			& 			&				& \checkmark		& \checkmark	& \checkmark \\
\hline
TSC	&			& 		& 		& 			&			&			& 		& 			& 			&				& 		& \checkmark	& \checkmark \\
\hline
Overhead & 0.55	& 0.56		& 0.81		& N/A				& 0.59		&  4.53		& 0.56		& 0.84		& 0.84		& 0.95	   		& 1.07		& 0.23	&  0.32 \\

\end{tabular}
\caption{Comparison of state-of-the-art self-checking circuit synthesis techniques with the evolved unconstrained ($E_U$) and non-intrusive ($E_{NI}$) methods we present.  H\&I refers to those using heuristics and iteration, NI to non-intrusive methods, FS to those which guarantee fault-secureness and overhead is presented as a fraction of duplication overhead averaged across benchmarks tackled by each publication. Since the benchmarks differ between publications, the overhead row is included merely to give a sense of performance. Rigorous comparisons for specific benchmarks are given later in the paper.}
\label{tablecompetition}
\end{table*}

A classic basic CED technique is duplication in which the output function generator and check symbol generator ($G$ and $C$ in Fig. \ref{fig2mod}) are identical and the checker simply compares their output. However, it is highly desirable to reduce overhead below that needed for duplication not only to lower production cost and power consumption but also to reduce mean time to failure. Hence CED design methods typically measure their performance in terms of percentage of duplication overhead, the lower the better -- figures of around 55\% are generally regarded as very good in most circumstances \cite{stan:03}.

However, it must be noted that in many cases in the literature full details of the original circuit synthesis procedure have not been published. In the case that it was suboptimal, then artificially low overhead figures will be quoted as checking overhead will be compared against a bloated duplication solution. For instance, in \cite{de:94} it is claimed CED is added whilst reducing total circuit area -- this could only be the case if the original circuit was suboptimal and a smaller implementation existed.  Thus  \cite{de:94} is excluded from Table \ref{tablecompetition}. Whenever the self-checking synthesis process presented in this paper reduces the output function generating logic beyond that synthesised by the thorough initial optimisation script (see \S\ref{appSis}) the resulting reduced circuit is used for duplication overhead comparison.

All previously reported TSC synthesis techniques fail to guarantee their checkers will be ST for circuits with limited output codespaces.  For instance, the standard duplication approach dual-rail checker tree fails to be fully exercised (and is thus not ST) in 6 out of 10 of the benchmarks tackled in this paper.  No previously presented technique addresses this issue automatically and hence \emph{there is no previously reported automatic end-to-end TSC circuit synthesis method}.  In general special care must be taken with modular designs to avoid such interfacing issues and ensure the overall circuit is TSC.  A further example is \cite{dali:13} in which the authors' claim of TSC generation in the case of irredundant original circuits requires their parity prediction logic being self-testing for which no evidence is provided.  Self-checking circuits using non-systematic codes have the hidden cost of requiring extra TSC code translators at the outputs.  Focusing the TSC property on the circuit as a whole, as in the method presented in this paper, avoids such interfacing issues.

Previous research \citep{sell:68,pies:02,bolc:97,bolc:00,toub:97} has suggested that the use of ad hoc codes and CED techniques to suit the particularities of a circuit can produce resource efficient TSC circuits.  Metra et al. \cite{metr:08} look at the information redundancy inherent in the target circuit in order to tailor a low overhead CED solution.  Indeed most recent self-checking research \cite{hong:11,mich:15} renounces general TSC synthesis for manually crafted ad hoc solutions for specific circuits.

Many successful CED synthesis techniques use heuristics and (at times stochastic) iteration.  Touba et al. \cite{toub:97} use a greedy search algorithm to improve parity grouping allocation using an estimated area cost function.  Dalirsani et al. \cite{dali:13} use SAT-based formal analysis to iteratively split an initial optimistic parity grouping until the circuit is fault secure (\S\ref{defnot}).  In fact heuristics and iteration was suggested as a design process by TSC circuit design pioneers. Carter et al. \cite{cart:68} state that ``a TSC computer can be designed using synthesis techniques based on iteration between tentative function designs and checker designs.  This iteration process is controlled by a probabilistic means for evaluating the effectiveness of the dynamic checking'', and Anderson \cite{ande:71} asserts:  "Small self-testing circuits are much easier to design by trial-and-error ".  Bouricius et al. \cite{bour:69} suggested a trial-and-error method of design using a program to simulate a proposed circuit design under failure conditions. 

Population based metaheuristic optimisation algorithms such as Brain Storm Optimisation \citep{shi:15} and Evolutionary Algorithms (EAs) \citep{holl:75,harv:92a,eiben:2010} are loosely inspired by biological concepts and use heuristics and iteration to search an encoded solution space by applying repeated selection and heritable variation to a population of tentative designs.  Thus through trial and error increasingly effective solutions are selected to populate subsequent generations.  This procedure matches the pioneers' vision of how to synthesise self-checking circuits.  Let $S$ be the set of solutions to a problem given an implementation technology, and $S_C$ the subset of $S$ using conventional design methods.  It is likely $S_C$ will include some of the better solutions in $S$ however it will also most likely miss many superior ones as well, especially considering that in most cases tailoring a solution to the particular instance of the problem will reap quality benefits.  In contrast, conventional design approaches attempt to apply a similar solution to every problem instance.  For instance TSC may be added with lower overhead to some circuits using parity checking, and to others through Berger codes.  Likewise for each circuit there is likely to be a specific approach that will be best, \emph{and this approach may be unique to that circuit}.  An EA is capable of efficiently finding solutions in $S_C$ and also exploring the better ones in $S \cap \overline{S_C}$ \footnote{In theory random search is also capable of discovering the solutions, but an EA does so much more efficiently \cite{holl:75}.}.  By operating on the circuit as a whole, at a fine-grained component level, it is not constrained to adopt the modularity and encapsulation required by standard techniques (Fig. \ref{fig2mod}).  In this paper we apply such an evolutionary process to the synthesis of TSC circuits.

Examples of the successful application of EAs to engineering (including hardware \cite{koza:2003,lohn:01,Tlelo-Cuautle:2010,trefzer:2017,vasi:14,mill:00c}) design are extensive \cite{gen:2007}.  There are several examples in the literature of evolved designs exploiting the particularities of the medium and problem instance to operate efficiently \cite{thom:97,miller:2014}.  The use of a Genetic Algorithm (GA), a form of EA, to synthesise small self-testing circuits is introduced in \cite{garv:03}.  Here we extend that work.

In contrast to the method introduced here, most approaches \cite{toub:97,stan:03,das:99,dali:13} dissuade logic sharing and optimisation to maintain TSC ability.  \emph{It is an open question wether TSC designs exist with a high level of function and checker logic sharing and optimisation}.  \emph{It is also an open question if a fully tailored efficient CED approach exists for every circuit and wether an automatic synthesis tool can generate it.}

In this paper we present the first automatic general synthesis method which can generate TSC circuits from arbitrary logic.  This is also the first synthesis method which tailors the self-checking strategy to the particularities of the given circuit whilst achieving very low overhead (an average of 23\% duplication across a set of widely used benchmark problems, significantly improving on previous best results for every problem in the set).  This is achieved without modular decomposition whilst sharing output function generator and checking logic.  The resulting circuits are TSC with respect to all gate input and output stuck-at faults.  The method does not affect input/output encoding.

We also present a variant of the technique which can be regarded as the first non-intrusive TSC synthesis method capable of adding logic around an existing (irredundant  \citep{arms:66}) circuit to make it TSC.  It also hass very low overhead (32\% of duplication for the same set of benchmarks).

Section \ref{defnot} will introduce some definitions, section \ref{combTSC} will present the synthesis method, section \ref{results} the results including analysis of evolved TSC circuits, and section \ref{digSCConc} will discuss conclusions.

\section{Definitions}
\label{defnot}
\label{digitalSCDef}
Given a circuit $G$ and a fault set $F$,

\emph{Definition 1}: $G$ meets the TSC goal (TSCG) \cite{smit:78} if it produces an error signal before or accompanying the first erroneous output due to a fault in $F$.

\emph{Definition 2}: $G$ is \emph{self-testing} (ST) if, for every fault in $F$, the circuit signals an error for at least one input codeword applied during normal operation.

 \emph{Definition 3}: $G$ is \emph{fault-secure} (FS) if, for every fault in $F$, the circuit never produces incorrect output without an accompanying error signal for input words applied during normal operation.

\emph{Definition 4}: $G$ is TSC with respect to $F$ if it is both ST and FS.  A TSC circuit will meet the TSCG under the assumption that all input codewords are applied between fault arrivals.

\emph{Definition 5}: A path is sensitised \citep{arms:66} from line $l_i$ to line $l_j$ if a change in value at $l_i$ will change the value at $l_j$.
\section{Synthesis Process}
\label{combTSC}
\label{combSynthProc}\label{evoDigitalSC}
The method proposed is a GA searching a space
of circuit designs.  Each design encountered is instantiated in a
digital logic simulator and evaluated on several metrics including
how ST and FS it is when subjected to faults.  These
fitness metrics influence the probability of a design being
selected for reproduction into the next generation.  The multiple
aspects of this process will now be described in greater depth.
Many of the choices described were arrived at after
experimentation with a significant number of inadequate
alternatives.  For this work, the logic gate library consists of
all possible two-input functions and the fault set includes
stuck-at faults at gate inputs and outputs.  These were chosen as
a standard for comparison yet the approach is general for any
gate library and fault set.

\subsection{Stochastic Optimisation} \label{ga} \label{digSCGA}
At the heart of our evolutionary approach are a number of interacting generational GAs
in the style of \cite{harv:92a}, each with
a fixed size population of 32 binary genotypes. These GAs are distributed over a large number of processors as described later. Each population is made up of
fixed length binary strings (genotypes) encoding a circuit as described in \S\ref{mapping}, and each circuit has a fitness measuring
its quality as a circuit with CED.  Once the fitness of every
individual in the population has been evaluated as in
\S\ref{evoCED}, they are sorted in descending order according to fitness. Members of the population are selected for reproduction with probability proportional to their fitness such that the fittest (top ranked) individual is twice as likely to be selected as the median ranked, with the remaining probabilities linearly interpolated. The two
highest ranking individuals are designated elites and are copied unmodified
into the next generation.  Six further individuals of the  next
generation are each formed by the sexual recombination of two
selected individuals through single point crossover: a random point $p$ is
picked and the offspring is formed by the genetic material of
parent $a$ up to $p$ and by that of parent $b$ after $p$. Another
16 are formed by mutating selected individuals with a single bit mutated (flipped)
at a random position in the genotype.  Two other new individuals are generated by a block-copy (translocation) mutation operation acting on selected population members which copies a section from one part of the genotype to another.  This is done so that a section representing one logic gate and its routing replaces that of another logic gate.  The remaining individuals
are formed by mutations affecting the encoded routing (see
\S\ref{mapping}) such that a mutation could change the source of a
gate to be the output of any other gate or a primary input.
This process of variation and selection 
leads to the fitness of the population increasing with subsequent
generations.

The GA style adopted is of a small genetically
converged population evolving for a large number of generations
with mutation -- including specially designed problem specific operators -- as the main driving force. Exchange of genetic material between the individual GAs, as described below, adds an additional powerful element to the search process.

\begin{figure}
\centering
\includegraphics[width=3.5cm]{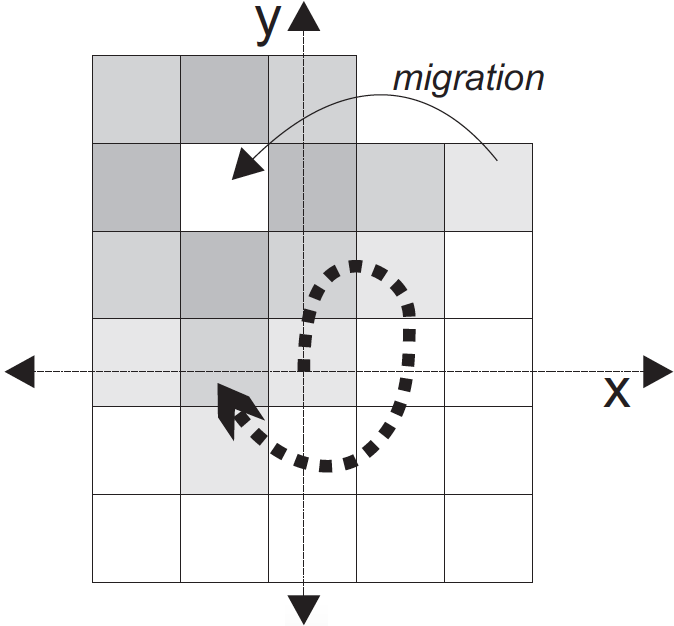}
\caption{Island based coevolutionary model: each square represents
an island. The arrow spiralling out from (0,0) delineates the order in which new islands are added
onto the grid. The shading intensity represents the probability of an island being selected as the
source of an immigrant to the island at (-1,2) with an example migration from (2,2).}
\label{figtop}
\end{figure}

Because it has been shown to be a highly effective style of EA \cite{Muhlenbein:89, Husbands:ec:94}, and in order to harness the idle processing power of many
workstations to aid the synthesis process, the overall search algorithm was distributed
using an island based coevolutionary model \citep{dhep}.  One
evolutionary process (GA)  as described in the previous paragraphs was
setup on each processor which was assigned a location on a 2D
grid (Fig. \ref{figtop}): an `island'. Individuals were selected for emigration between islands using the same
scheme as for reproduction (\S\ref{ga}).  Their destination
was random such that the probability of choosing an island varied
inversely with its distance to the migrant's source.  The
probability of each migration event succeeding was configured to
maintain genetic diversity across populations yet allow the
diffusion of fit individuals. A grid of up to 150 workstations was used. The distributed nature of the algorithm meant that it could seamlessly handle a dynamic number of islands as processor availability came and went.  As all islands are seeded (\S\ref{seeding}) with an implementation of the original circuit together with random bits, the whole grid begins with an `exploring' stage with a sea of mostly random genetic material.  As migration spreads fitter individuals, the grid enters an `exploiting' stage with greater convergence.

\subsection{Genotype to Circuit Mapping} \label{mapping} \label{combMap}
Every binary genotype encodes a circuit structure which is
instantiated during fitness evaluation.  During evolutionary
synthesis circuits are constrained to have $q$ outputs, $r$
inputs and a maximum of $2^{b}-r$ two-input logic gates, where
$b$ bits are used for a routing address.  The first $q \times b$
bits define what gates will drive primary outputs.  The rest of
the genotype is formed by $2^b-r$ genes of $4+2 \times b$ bits,
each gene defining the gate whose address is its position in the
gene list.  The $r$ largest addresses point to primary inputs.
The first four bits within a gene are the truth table for the
represented two-input gate, while the remaining $2\times b$ bits
encode the addresses of its inputs.  Each of these $b$ bit blocks
may be affected as a whole by the routing mutations described in
\S\ref{ga}.  $b$ is fixed manually per run to be the smallest value allowing a duplication approach to be encoded.

\label{mappingFF}This mapping allows for the encoding of circuits with recurrency.  However, genotypes were forced to be feed-forward as follows: a depth-first
search is performed on the encoded circuit starting from each
output in turn.  If during the search of a single path a gate is
encountered for a second time, then the loop is broken by
re-routing to a random primary input instead.  Even though a maximum number of
gates is allowed, any gates which do not feed POs (including
error rails), and hence play no functional role, are discarded during simulation. Thus the mapping can effectively encode a variable number of gates. This mapping is
more flexible and defines a fitness landscape \cite{Smith:2002} with richer
connectivity than the standard feed-forward mapping used for
hardware evolution \citep{vass:00a}. 

\subsection{Circuit Simulation} \label{digSCSim}
In order to evaluate their fitness as self-checking circuits, the
genotypes are instantiated in a simulator, their structure
determined by the mapping described above.  The simulator used is
our own carefully verified event driven digital logic simulator in
which each logic gate is in charge of its own behaviour when
given discrete time-slices and the state of its inputs.  Logic
gates may perform any two-input function.  Gates are sorted such that no gate in the list
drives an input to a gate at an earlier position and gates are refreshed in a single wave in
order.   The fault set $F$ used in this publication is of all gate input and output stuck-at faults.  A stuck-at fault at a wire is simulated simply by fixing it to 0 or 1.

\subsection{Primary Function Evaluation} \label{evoOFG}
Once a genotype is instantiated as a circuit in the simulator,
the evaluation of several qualities is ready to begin.  This
section describes the procedure used to evaluate a circuit's
ability at generating the correct output function.  All $2^{r}$
input words are applied once \footnote{A limited input codeword space could be used when appropriate.}.  Let
$Q_j$ be the concatenated response at output $j$ sampled as
described in \S\ref{digSCSim} to all input words applied, and $Q'_j$
be the desired response.  Then output function fitness $f_f$ is
the modulus of the correlation of $Q_j$ with $Q'_j$ averaged over
all $q$ function outputs:

\begin{equation}\label{ff}
  f_f = \sum_{j=0}^{q-1} \frac{|corr(Q_j,Q'_j)|}{q}
\end{equation}

Allowing inverted outputs improves the smoothness and
connectivity of the fitness landscape while it remains trivial to
change an evolved circuit's outputs to be non-inverting if
required.  A circuit with $f_f=1$ generates the output function
correctly.

\subsection{Evaluation of Self-Checking Ability} \label{evoCED}
In order to gauge a circuits' quality at CED, the behaviour of the
extra outputs forming the dual rail error signal $z_0z_1$ is
observed in the presence of injected faults.  Fitness metrics
$f_{ST}$ and $f_{FS}$ measure the degree to which a circuit
fulfills the ST and FS criteria.  If an error is signalled -- ie.
$z_0=z_1$ -- during fault-free evaluation of $f_f$ then
$f_{ST}=f_{FS}=0$.  Otherwise the same evaluation procedure
described in \S\ref{evoOFG} is repeated under every fault in the set of gate output faults $F_q$, and $f_{ST}$ and $f_{FS}$ are calculated as follows:

$f_{ST} = \frac{1}{1+u_f k_{ST}}$ where $u_f$ is the number of
faults for which an error was never signalled, and $k_{ST}$ was
chosen to be 25.

$f_{FS} = \frac{1}{1+u_i k_{FS}}$ where $u_i$ is the number of
fault-input word instances for which output was incorrect and an
error was not signalled, and $k_{FS}$ was chosen to be 200.

Constants $k_{ST}$ and $k_{FS}$ were chosen to give $f_{ST}$ and
$f_{FS}$ good sensitivity when $u_f$ and $u_i$ are small.  It is
now demonstrated how a circuit which achieves maximum
fitness under these metrics is TSC:

\begin{theorem}
\label{thevalSCLab}
If a circuit $G$ has $f_f=f_{ST}=f_{FS}=1$ when
evaluated with fault set $F_x$, then $G$ is TSC with
respect to $F_x$.
\end{theorem}
\begin{IEEEproof}
Since $f_f=1$ and $f_{ST}>0$ then the output
function is correct and no errors are signalled during fault-free
operation.  Since $f_{ST}=1$ then $u_f=0$ and there is no fault
for which an error is not signalled during normal operation.
Hence $G$ is ST by Definition 2.  Since $f_{FS}=1$ then $u_i=0$
and there is no fault/input word combination for which an incorrect output is not accompanied by an error signal.  Hence $G$ is FS by
Definition 3.  Thus $G$ is TSC by Definition 4.
\end{IEEEproof}

In order to reduce computational effort required for evaluating
self-checking ability the fault set $F_q$ used only includes
faults at gate outputs.  However,
\begin{theorem}
\label{thFSLab}
If a circuit $G$ is FS with respect to all
gate output faults $F_q$ then it is also FS with respect to all
gate input and output faults $F$. 
\end{theorem}
\begin{IEEEproof}
A stuck-at fault $f$ at an input of a gate $g$
will, depending on circuit input word $x$, either invert the
output of $g$ or leave it unchanged.  If unchanged then circuit
output is unchanged and correct.  If the output of $g$ is
inverted then $f$ manifests itself as a gate output fault at $g$ stuck-at the inverted value.  Since
$G$ is FS with respect to $F_q$ then incorrect circuit output
will not be produced without an accompanying error signal.  Since
this holds for an arbitrary fault $f$, gate $g$ and input $x$,
then $G$ is FS with respect to all faults at gate inputs.
\end{IEEEproof}
\label{combGIF}
Unfortunately this does not hold for ST because the full
exercising of every gate during normal operation is not
guaranteed for redundant circuits.  Even though circuits under parsimony selection pressure (see next section) are increasingly unlikely to be redundant, the ST property of gate input faults was measured during simulation of gate output faults as follows.  A gate input fault will generate an error signal under an input word if and only if it manifests itself as a gate output fault generating an error signal.  Whenever a gate output fault generates an error signal under some input word during evaluation, the gate input faults that would have manifested themselves as that gate output fault are recorded.  This is done by removing the gate output fault and recording which gate input inversions restore the faulty output value.  All gate input faults which never manifest themselves as gate output faults signalling an error were counted in $u_F$ when calculating $f_{ST}$.  Thus all
combinational results presented -- with $f_{ST}=f_{FS}=1$ -- are circuits with TSC CED with
respect to all faults at gate inputs and outputs.

\subsection{Combined Fitness and Parsimony} \label{combpars}\label{SCPars}
The evaluation procedures above generate fitness metrics
measuring a circuit's quality with respect to generating the output function
($f_f$) and CED ($f_{ST}$,$f_{FS}$).  A final metric $f_p$
measures the size $s$ of a circuit such that $f_p$ varies from 0
(maximum size) to 1 (size 0) as $f_p=\frac{M-s}{M}$ where maximum
size $M=2^b-r$ (see \S\ref{mapping}).  Gates with no downstream path to circuit outputs are excluded from $s$.  No functional circuit will
have $f_p=1$, yet since the smallest possible implementation of a
circuit with TSC CED is not known, the evolutionary synthesis
process was allowed to increase $f_p$ as far as possible within
the practical constraints imposed by processing power available and was stopped manually after a maximum of two weeks.

Even though optimisation of all fitness metrics is desirable,
evolution is encouraged to prioritise generating the correct
output function above performing self-checking, and the latter
above minimising overhead.  This is achieved by defining a multi-objective
fitness vector $(f_f,f_{ST},f_{FS},f_p)$ and sorting the
population for rank selection according to the dictionary total
ordering defined, such that later metrics are only compared when
earlier ones are equal.  
The effect of this is to automatically shift the main selection pressure to the next fitness function as soon as previous ones have been fully optimised, thus defining an incremental evolution \citep{harv:94} path.  Even though at any moment there will be only one unoptimised metric with highest priority driving selection, when this metric is equal for two individuals the GA will ``look ahead'' and use the next (lower) priority metric to judge which individual is best.  This is rather like a driver attending to immediate demands but at the same time keeping in mind her long term destination.

\subsection{Seeding and Locking the Original Circuit} \label{oofg} \label{seeding}
All evolutionary runs were initialised with a population of individuals encoding an
output function generator circuit synthesised by Sis
\citep{sent:92} using the procedure described in \S\ref{appSis}. In the initial population the parts of the genetic material of each individual not involved in
the encoding of this circuit were randomly generated.  For non-intrusive runs, this original
output function generator circuit was fixed so that no genetic operator could modify it.

\section{Results}
\label{results}
\subsection{Overhead Comparison}
All circuits presented in this section achieve $f_f=f_{ST}=f_{FS}=1$ when
evaluated with the fault set including stuck-at gate inputs and
outputs and therefore perform TSC CED by Theorem 1.  Further
proof for one example will be presented during the analysis of circuit A in \S\ref{HotBrain}.

Circuits synthesised with the proposed method were compared to
equivalents in the literature in terms of checking overhead as a fraction of duplication and comparison overhead.  Comparisons were made with duplicated circuits synthesised using the
recommended Xilinx scripts for LUT technology using 
Sis (see \S\ref{appSis}).  This ensured a fair comparison since these tools
exploit the full range of gates available.  Choosing an unconstrained gate set allows
analysis of what kind of gate the evolutionary process finds
useful for building self-checking circuits.  Hardware cost is measured in this work in terms of gate count,
which, like the literal count or gate area used in many previous
publications, ignores area required for routing. When other techniques present overhead in terms of area including routing and literal count their literal count figure will be quoted as it is more similar to gate count.

\subsection{The Benchmarks}

\begin{table*}
\scriptsize
  \centering
  \caption{Overhead comparison of unconstrained evolved $E_U$, non-intrusive evolved $E_{NI}$
and previous best PB in the literature.  All evolved and \citep{efan:16} circuits are TSC.  \citep{goes:00,moro:00} are partially self-checking, \citep{toub:97} are FS.}
\label{tabOH}
  \begin{tabular}{|C{1.5cm}||c|c|c|c||c|c|c||c|c|c|}
    \hline
    \multirow{2}{1.5cm}{\centering Benchmark}&\multicolumn{4}{c||}{Original}&\multicolumn{3}{c||}{Oh. gate count} &\multicolumn{3}{c|}{Oh./Dup. Oh.} \\
    \cline{2-11}
    &Ins. & Outs. & Lits. & Gates & $E_U$ & $E_{NI}$ & Dup. & $E_U$ & $E_{NI}$ & PB \\
    \hline
Mult2&4&4&28&7&  8&8&25&0.32&0.32&\\
b1&3&4&17&5  & 3&5&23&0.13&0.22&0.74 \citep{goes:00}\\
c17&5&2&12&6 &7&9&12&0.58&0.75&\\
cm82a&5&3&55&10& 5&9&22&0.22&0.41&0.96 \cite{efan:16}\\
cm42a&4&10&35&18 &8&12&72&0.11&0.17&0.75 \citep{moro:00}\\
wim&4&7&103&22 & 8&-&58&0.14&-&0.27 \citep{toub:97}\\
dc1&4&7&56&29& 17&-&65&0.26&-&0.53 \citep{toub:97}\\
cm138a&6&8&35&16  & 9&10&58&0.16&0.17&0.81 \citep{moro:00} \\
rd53&5&3&182&23& 5&-&29&0.17&-&0.82 \cite{efan:16}\\
decod&5&16&68&26   &16&20&116&0.14&0.17&0.60 \citep{goes:00} \\
rd73&7&3&741&49 & 6&-&47&0.13&-&0.74 \cite{efan:16}\\
m1&6&12&233&64&52&-&194&0.4&-&\\
\hline
Average&&&&&&&&0.23&0.32&0.69\\
\hline
  \end{tabular}
\end{table*}

In order to provide a meaningful evaluation of the proposed
technique, circuits belonging to the MCNC'91\cite{yang:91} combinational suite appearing most frequently in previous papers were
adopted as benchmarks\footnote{A small two bit multiplier is also included for reference.}.  

\subsection{Synthesised TSC Circuits}
The computational cost for the synthesis of the
circuits presented varied from a couple of hours
on a single 2.8Ghz processor to a couple of weeks on 150 2.8Ghz
processors.

The evolutionary synthesis process described in \S\ref{combSynthProc}
arrived at circuits with TSC CED for all benchmarks for which the
computational cost of fitness evaluation did not render the
approach impractical.  Circuits with TSC CED were synthesised for
each benchmark both allowing modifications to the original output
function generator circuit and non-intrusively. In the latter case, checking logic was appended around the existing design such that
the only structural change to the output function generator was
the extra routing required to feed the checking logic.  

Table \ref{tabOH} compares the overhead of evolved circuits with
TSC CED to that required for duplication.  Duplication overhead
includes one extra copy of the original circuit (with all outputs
inverted at no extra cost) and the required amount of two-rail
checkers to check all outputs and produce a single two-rail error
signal.  Each two-rail checker requires 6 gates \cite{hugh:84}, so for a circuit
with $g$ gates and $q$ outputs duplication overhead amounts to
$g+6(q-1)$ gates.  It is important to note that in many cases such a standard duplication solution will require extra logic to fully exercise the checker under the circuit output codespace.  Hence our approaches' overheads are actually even lower compared to truly TSC duplication solutions.  Table \ref{tabOH} then compares the overhead required as a
fraction of duplication overhead for the proposed locked and
unconstrained evolutionary techniques and for previous techniques
found in the literature.  Only in the cases of rd53 and rd73 did evolution optimise the output function generator logic beyond the version synthesised by Sis.  This smaller output function generator is then used to calculate duplication overhead instead of the larger one produced by Sis.  It is impossible to know when a circuit is maximally optimised \citep{gunt:99} but the fact that for all other circuits evolution could not beat Sis shows that the latter was reasonably good.  This means the overhead figures for this method are likely even lower in comparison to other methods which are not capable of such extra optimisations of the function logic which would especially come into play as circuit size grows.  This is the case because evolutionary search is also capable of finding minimal implementations of circuits unreachable by conventional methods \cite{vasi:14}. 

For all benchmarks attempted, circuits produced with the proposed
unconstrained approach require less overhead as a fraction of
duplication overhead than all previously published techniques
\citep{jha:93,de:94,toub:97,das:99,moro:00,goes:00} by a
considerable margin. This is true even for all
of those circuits for which the original output function
generator design was not modified. It must be noted that these
circuits are by no means the best evolution could find since
informal tests showed that smaller solutions were found when runs were left for longer periods -- indeed circuit A (\S\ref{HotBrain}) was superseded by an even smaller overhead solution close to publication. Since practical time limits were placed on the runs featured in the table (a maximum of two weeks running the distributed evolutionary algorithm), it is an open question as to what the theoretical minimum
overhead is for TSC CED and if this method can find it.  Table \ref{tabOH} also includes comparisons of synthesis computational cost.  Eval. $\Phi$ denotes circuit evaluation effort, Evals. the number of evaluations per run, and $\Sigma \Phi$ shows total run effort.  Overall processing time is dominated ($>99.8\%$) by circuit evaluation time.  The genotype to phenotype mapping time is minimal.  Unconstrainted evolved
circuits required on average 23\% of the overhead required by the duplication approach.  This was equivalent to on average an overhead of 63\% of the logic used for the original function generating logic.  Circuits evolved with no modifications to the original design required on
average 32\% duplication overhead, or 83\% overhead with respect to the
locked function generator.  Such low overhead figures are unheard of in the literature.  Immediately several questions spring to mind.  How can it be that evolved designs requires so little
overhead?  How do they operate?

\begin{figure*}
\centering
\includegraphics[width=12cm]{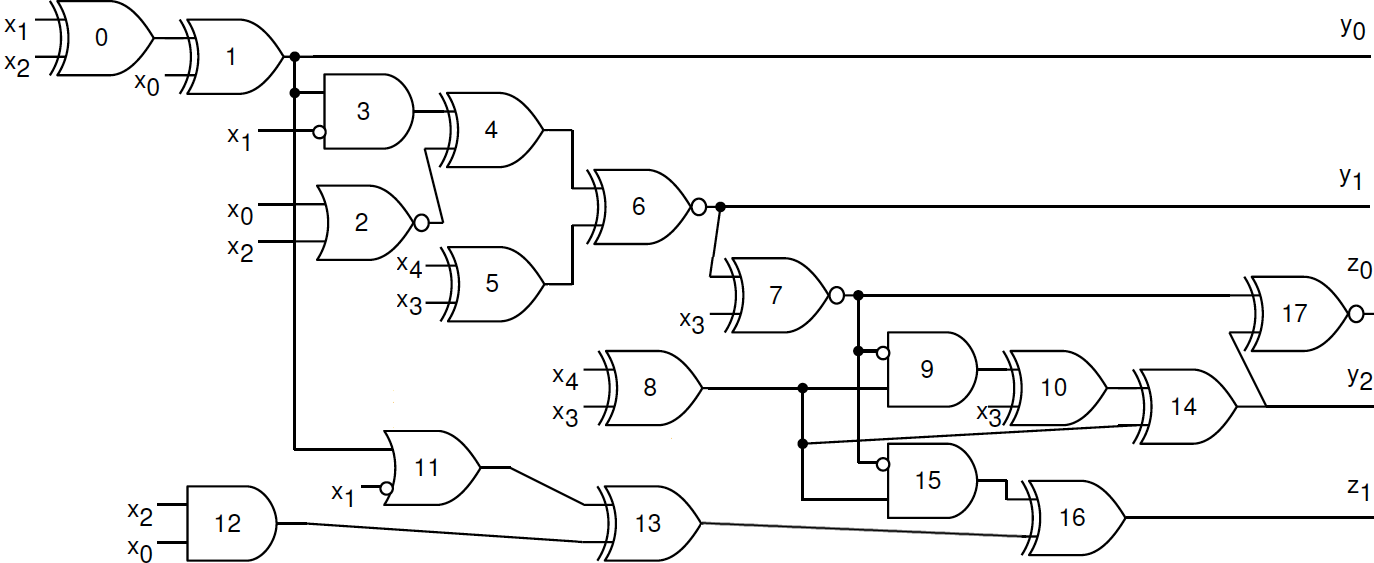}
\caption{Circuit B: non-modular TSC cm82a benchmark with 4 gates overhead
instead of the 26 required for duplication with comparison.  The original circuit has been modified by the synthesis process such that many gates generate both POs and error signals.} \label{figRocio}
\end{figure*}

Simple inspection of evolved designs such as that in Fig.
\ref{figRocio} reveals their structure is not constrained to
the function-checker model of Fig. \ref{fig2mod}.  Thus \emph{the proposed method
becomes the first automatic synthesis technique to generate TSC
circuits not adopting the function-checker
decomposition}.  This opens up a whole class of circuits with TSC
CED to the evolutionary design process, some previously discussed
in the literature \citep{sell:68,moha:03b}.  Belonging to this
class are circuits in which the
function-checker distinction is blurred and individual gates are both
used for calculating POs and controlling the
propagation of error signals.  Also belonging to this class are
circuits with dependency between the outputs such that only some
need be checked, or circuits where the function-checker is
roughly adopted but the checker checks internal lines as well as
outputs\footnote{Something similar is done in \citep{moha:03b} for partially
self-checking circuits.}.  Clearly not all circuits belonging to
this non-modular class will have lower checking overhead than those with the
function-checker decomposition, but the properties mentioned above
explain how many of them do.  The operating principles behind
evolved circuits with TSC CED and arbitrary structure will be
presented at greater depth in the following sections.

Further inspection of evolved circuits reveals that they use
multiple strategies to achieve TSC CED behaviour, across
different circuits and even within the same one.  For example
the circuit in Fig.\ref{figHotBrain} uses a strategy akin to parity group checking
whilst circuit B (Fig.\ref{figRocio}) makes heavy use of output cascading and error propagation.  The strategies used by
evolved designs are not random but seem to be well adapted to the
particularities of the original circuit.  Thus \emph{the proposed
technique becomes the first capable of synthesising circuits with
TSC CED adopting multiple strategies suitable to the
particularities of the circuit at hand}.  As shall be
demonstrated below, some of the strategies employed by evolved
designs are to the authors' knowledge novel, and efficiently
tailored for the circuit at hand.  Thus \emph{the proposed
technique is capable of creating novel ad hoc strategies to
achieve TSC CED exploiting the particularities of the original
circuit}.  The
advantage of using ad hoc TSC strategies for particular circuits
have been previously discussed
\citep{sell:68,pies:02,bolc:97,bolc:00} but this is the first
automatic synthesis method to design them.  This is also \emph{the first method to make an existing circuit TSC non-intrusively}.

\subsection{Analysis of synthesised design}
\label{HotBrain}
This section will present circuit A: an evolved TSC cm42a benchmark with 14\% duplication overhead.  Circuit A (Fig. \ref{figHotBrain}) performs the output function defined by the cm42a benchmark with 4 inputs and 10 outputs and has been synthesised
using the process described in \S\ref{combSynthProc}
allowing modifications to the original seeded output function
generator logic.  This logic was itself synthesised by Sis and
used 18 gates.  Gate addresses were chosen to be represented by 6
bits allowing a maximum circuit size of 60 gates during
synthesis. CED using duplication would require 18 gates for the
copy and 54 for the checker totalling 72 gates overhead.  Circuit A has 10 gates of checking overhead.  The maximum delay from primary inputs to outputs increased from 3 to 5 gate delays and the maximum error latency is 4 gate delays.

\subsubsection{Proof Circuit A is TSC}
Circuit A has already been proved to be TSC by Theorem \ref{thevalSCLab} because it scores $f=f_{ST}=f_{FS}=1$.  Additional automated and manual proofs will be laid out below and the reader is also invited to tinker with a spreadsheet implementation at  {\footnotesize \texttt{http://dhep.ga/files/circuita.xls}} and a visually simulated version at {\footnotesize \texttt{http://dhep.ga/circuitviewer.php?ind\_id=3466}}.

\emph{Normal Operation:}
Table \ref{hotbrainfaultfree} proves circuit A performs the cm42a function correctly (with inverted\footnote{Inversion is allowed (see Equation \ref{ff}) as it is cost-free for any downstream module in 2-input-LUT technology.} output $y_3$) and $z_0 \neq z_1$ during fault-free operation.  The latter is also evidenced by the simplification of their Boolean equations extracted from the circuit using the Quine McCluskey algorithm: $z_0 = \overline{x_0}+\overline{x_1}+x_2+x_3= \overline{ x_0 x_1 \overline{x_2} \overline{x_3}}$ and $z_1=x_0 x_1 \overline{x_2} \overline{x_3}$.  This algorithm was also used to independently (and diversely) confirm the ST and FS properties by replacing faulty units with 0 or 1 in the Boolean expressions for $y_i$ under every input word/fault combination, thus providing a second automated proof that circuit A is TSC.  A third manual proof is laid out as follows:
\setlength\tabcolsep{2pt}
\begin{table}
\centering
\caption{Fault-free operation of circuit A.  Gate $g_i$ indices match Fig. \ref{figHotBrain}.  }
\label{hotbrainfaultfree}
\begin{scriptsize}
\begin{tabular}{|c|c							c	c	c	c	c	c	c	c									|c	c	|c	c|
}
\hline
$$&						$y_5$&							$y_4$&	$y_0$&	$y_1$&	$y_8$&	$y_2$&	$y_7$&	$y_9$&	$y_6$&								$$&	$$&	$z_0$&	$y_3{=}z_1$	\\
$x$&						$g_2$&							$g_9$&	$g_{10}$&	$g_{11}$&	$g_{12}$&	$g_{13}$&	$g_{14}$&	$g_{15}$&	$g_{16}$&								$g_{24}$&	$g_{25}$&	$g_{26}$&	$g_{27}$	\\
\hline
0000&						1&							1&	0&	1&	1&	1&	1&	1&	1&								1&	0&	1&	0	\\
0001&						1&							1&	1&	1&	0&	1&	1&	1&	1&								1&	0&	1&	0	\\
0010&						1&							0&	1&	1&	1&	1&	1&	1&	1&								1&	0&	1&	0	\\
0011&						1&							1&	1&	1&	1&	1&	1&	1&	1&								0&	1&	1&	0	\\
0100&						1&							1&	1&	1&	1&	0&	1&	1&	1&								1&	0&	1&	0	\\
0101&						1&							1&	1&	1&	1&	1&	1&	1&	1&								0&	1&	1&	0	\\
0110&						1&							1&	1&	1&	1&	1&	1&	1&	0&								1&	0&	1&	0	\\
0111&						1&							1&	1&	1&	1&	1&	1&	1&	1&								0&	1&	1&	0	\\
1000&						1&							1&	1&	0&	1&	1&	1&	1&	1&								1&	0&	1&	0	\\
1001&						1&							1&	1&	1&	1&	1&	1&	0&	1&								1&	0&	1&	0	\\
1010&						0&							1&	1&	1&	1&	1&	1&	1&	1&								0&	1&	1&	0	\\
1011&						1&							1&	1&	1&	1&	1&	1&	1&	1&								0&	1&	1&	0	\\
1100&						1&							1&	1&	1&	1&	1&	1&	1&	1&								1&	1&	0&	1	\\
1101&						1&							1&	1&	1&	1&	1&	1&	1&	1&								0&	1&	1&	0	\\
1110&						1&							1&	1&	1&	1&	1&	0&	1&	1&								1&	0&	1&	0	\\
1111&						1&							1&	1&	1&	1&	1&	1&	1&	1&								0&	1&	1&	0	\\

\hline
\end{tabular}
\end{scriptsize}
\end{table}

\begin{figure*}
\centering
\includegraphics[width=11cm]{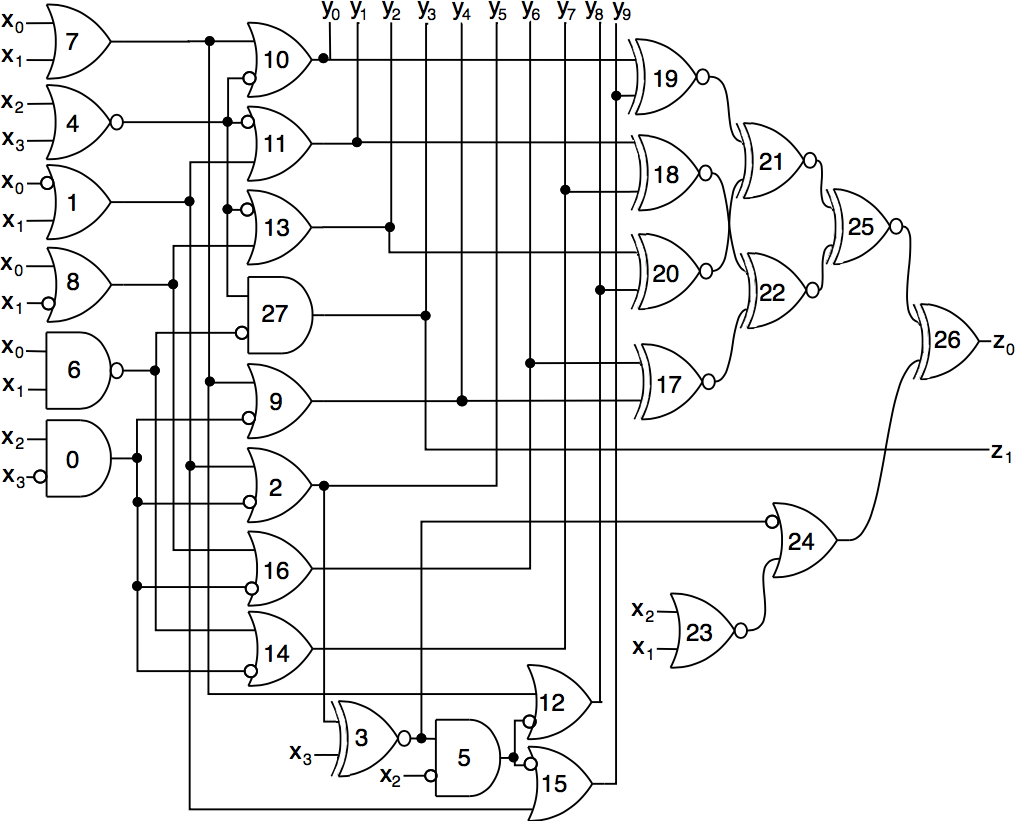}
\caption{Circuit A: evolved TSC cm42a benchmark using 10 gates overhead
instead of the 72 required for duplication with comparison.} \label{figHotBrain}
\end{figure*}

\emph{Self-Testing:}
Let $a_i.d$, $b_i.d$, $q_i.d$ represent stuck-at-$d$ faults at the first input, second input, and output lines of gate $i$ respectively.  Gate indices are as per Fig. \ref{figHotBrain}.  An input word setting $z_0=z_1$ for each fault is listed:

\begin{scriptsize}
$a_0.0$: 0010, $b_0.0$: 0011, $q_0.0$: 0010, $a_0.1$: 0000, $b_0.1$: 0010, $q_0.1$: 0000, 
$a_1.0$: 1000, $b_1.0$: 1100, $q_1.0$: 0000, $a_1.1$: 0000, $b_1.1$: 1000, $q_1.1$: 1000, 
$a_2.0$: 0010, $b_2.0$: 1010, $q_2.0$: 0000, $a_2.1$: 1010, $b_2.1$: 1000, $q_2.1$: 1010, 
$a_3.0$: 0000, $b_3.0$: 0001, $q_3.0$: 0001, $a_3.1$: 1010, $b_3.1$: 0000, $q_3.1$: 0000, 
$a_4.0$: 0010, $b_4.0$: 0001, $q_4.0$: 0000, $a_4.1$: 0000, $b_4.1$: 0000, $q_4.1$: 0001, 
$a_5.0$: 0001, $b_5.0$: 0011, $q_5.0$: 0001, $a_5.1$: 0000, $b_5.1$: 0001, $q_5.1$: 0000, 
$a_6.0$: 1100, $b_6.0$: 1100, $q_6.0$: 0000, $a_6.1$: 0100, $b_6.1$: 1000, $q_6.1$: 1100, 
$a_7.0$: 1000, $b_7.0$: 0100, $q_7.0$: 0100, $a_7.1$: 0000, $b_7.1$: 0000, $q_7.1$: 0000, 
$a_8.0$: 1100, $b_8.0$: 0100, $q_8.0$: 0000, $a_8.1$: 0100, $b_8.1$: 0000, $q_8.1$: 0100, 
$a_9.0$: 0110, $b_9.0$: 0010, $q_9.0$: 0000, $a_9.1$: 0010, $b_9.1$: 0000, $q_9.1$: 0010, 
$a_{10}.0$: 0100, $b_{10}.0$: 0000, $q_{10}.0$: 0001, $a_{10}.1$: 0000, $b_{10}.1$: 0001, $q_{10}.1$: 0000, 
$a_{11}.0$: 0000, $b_{11}.0$: 1000, $q_{11}.0$: 0000, $a_{11}.1$: 1000, $b_{11}.1$: 1001, $q_{11}.1$: 1000, 
$a_{12}.0$: 0101, $b_{12}.0$: 0001, $q_{12}.0$: 0000, $a_{12}.1$: 0001, $b_{12}.1$: 0000, $q_{12}.1$: 0001, 
$a_{13}.0$: 0000, $b_{13}.0$: 0100, $q_{13}.0$: 0000, $a_{13}.1$: 0100, $b_{13}.1$: 0101, $q_{13}.1$: 0100, 
$a_{14}.0$: 0010, $b_{14}.0$: 1110, $q_{14}.0$: 0000, $a_{14}.1$: 1110, $b_{14}.1$: 1100, $q_{14}.1$: 1110, 
$a_{15}.0$: 0001, $b_{15}.0$: 1001, $q_{15}.0$: 0000, $a_{15}.1$: 1001, $b_{15}.1$: 1000, $q_{15}.1$: 1001, 
$a_{16}.0$: 0010, $b_{16}.0$: 0110, $q_{16}.0$: 0000, $a_{16}.1$: 0110, $b_{16}.1$: 0100, $q_{16}.1$: 0110, 
$a_{17}.0$: 0000, $b_{17}.0$: 0000, $q_{17}.0$: 0000, $a_{17}.1$: 0110, $b_{17}.1$: 0010, $q_{17}.1$: 0010, 
$a_{18}.0$: 0000, $b_{18}.0$: 0000, $q_{18}.0$: 0000, $a_{18}.1$: 1000, $b_{18}.1$: 1110, $q_{18}.1$: 1000, 
$a_{19}.0$: 0001, $b_{19}.0$: 0000, $q_{19}.0$: 0001, $a_{19}.1$: 0000, $b_{19}.1$: 1001, $q_{19}.1$: 0000, 
$a_{20}.0$: 0000, $b_{20}.0$: 0000, $q_{20}.0$: 0000, $a_{20}.1$: 0100, $b_{20}.1$: 0001, $q_{20}.1$: 0100, 
$a_{21}.0$: 0001, $b_{21}.0$: 0000, $q_{21}.0$: 0010, $a_{21}.1$: 0000, $b_{21}.1$: 0001, $q_{21}.1$: 0000, 
$a_{22}.0$: 0000, $b_{22}.0$: 0000, $q_{22}.0$: 0000, $a_{22}.1$: 1000, $b_{22}.1$: 0010, $q_{22}.1$: 0010, 
$a_{23}.0$: 0011, $b_{23}.0$: 0101, $q_{23}.0$: 0001, $a_{23}.1$: 0001, $b_{23}.1$: 0001, $q_{23}.1$: 0011, 
$a_{24}.0$: 0011, $b_{24}.0$: 0001, $q_{24}.0$: 0000, $a_{24}.1$: 0010, $b_{24}.1$: 0011, $q_{24}.1$: 0011, 
$a_{25}.0$: 0010, $b_{25}.0$: 0000, $q_{25}.0$: 0011, $a_{25}.1$: 0000, $b_{25}.1$: 0010, $q_{25}.1$: 0000, 
$a_{26}.0$: 0011, $b_{26}.0$: 0000, $q_{26}.0$: 0000, $a_{26}.1$: 0000, $b_{26}.1$: 0011, $q_{26}.1$: 1100, 
$a_{27}.0$: 1100, $b_{27}.0$: 0000, $q_{27}.0$: 1100, $a_{27}.1$: 1101, $b_{27}.1$: 1100, $q_{27}.1$: 0000, 
  
\end{scriptsize}

\emph{Fault Secure:}
Gates $G_y={ g_{9-16},g_{27} }$ can easily be seen to each only affect a single function output and any change in them will always change either $z_0$ through the NXOR tree or $z_1$ in the case of $g_{27}=y_3=z_1$.  It is then evident that any errors caused by faults at these maintain the FS property.  Gates $G_z={g_{17-26}}$ do not affect function output and hence fault secureness is maintained when they fail.

For the remaining nine gates, the full list of fault/input word combinations and resulting circuit outputs would require 288 rows and 10 columns of data -- as produced and verified during $f_{FS}$ calculation -- so we will attempt to summarise.  Table \ref{hotbrainFS} lists line/input word combinations and their corresponding downstream sensitised paths.  If the paths include any PO gate then they must also include either $g_{26}=z_0$ or $g_{27}=z_1$ for fault secureness to be maintained.  Since any single sensitising path to $g_y \in G_y$ always affects either $z_0$ or $z_1$ the whole path will be omitted for brevity after the first rows.  Input words for which the line $q_i$ (output of gate $i$) has no impact on POs are not listed.  This table is thus equivalent to a list of gate/input word combinations for which gate output errors will propagate to PO errors.  For each combination the table will also list POs affected and which error rail will flag the error.  Since gate input faults can only manifest themselves as gate output errors, this is sufficient to prove the circuit is FS with respect to all faults at these gates (Theorem \ref{thFSLab}).  Since only one error rail is affected by all error generating fault/input word combinations then all incorrect output has an accompanying error signalled and fault secureness is maintained.

\setlength\tabcolsep{2pt}
\begin{table}
\centering
\caption{Proof that circuit A is FS: fault/input word combinations impacting POs always flip a single error rail, because sensitised paths from gates to POs $y$ always continue to either $z_0$ or $z_1$.}
\label{hotbrainFS}
\begin{scriptsize}
\begin{tabular}{|C{0.65cm}|C{0.65cm}|c|c|c|}
\hline
Fault at & With Input & Affects (downstream sensitised path to) & $y$ & $z$ \\
\hline
$q_0$ & 00xx & $g_9 (y_4) \Rightarrow g_{17} \Rightarrow g_{22} \Rightarrow g_{25} \Rightarrow g_{26} (z_0)$ & $y_4$ & $z_0$ \\
   & 01xx & $g_{16} (y_6) \Rightarrow \cdots \Rightarrow z_0$ & $y_6$ & $z_0$ \\
   & 100x & $g_2 (y_5) \Rightarrow g_3 \Rightarrow g_5 \Rightarrow g_{15} (y_9) {\Rightarrow} {\cdots} {\Rightarrow} z_0$ & $y5, y9$ & $z_0$ \\
   & 101x & $g_2 \Rightarrow g_3 \Rightarrow g_{24} \Rightarrow z_0$ & $y_5 $& $z_0$ \\
   & 11xx & $g_{14} (y_7) \Rightarrow \cdots \Rightarrow z_0$ & $y_7$ & $z_0$ \\
$q_1$  & xx00 & $g_{11} (y_1) \Rightarrow \cdots \Rightarrow z_0$ & $y_1$ & $z_0$ \\
   & xx01 & $g_{15} \Rightarrow \cdots \Rightarrow z_0$ & $y_9$ & $z_0$ \\ 
   & xx10 & $g_2 \Rightarrow g_3 \Rightarrow g_{24} \Rightarrow z_0 $ & $y_5$ & $z_0$ \\
$q_2$  & 000x & $g_{12} (y_8) \Rightarrow \cdots \Rightarrow z_0$ & $y_5, y_8$ & $z_0$ \\
   & xx1x & $g_3 \Rightarrow g_{24} \Rightarrow z_0$ & $y_5$ & $z_0$ \\
   & rest & $g_3 \Rightarrow g_5 \Rightarrow g_{24} \Rightarrow z_0$ & $y_5$ & $z_0$ \\
$q_3$  & 000x & $g_5 \Rightarrow g_{12} \Rightarrow \cdots \Rightarrow z_0$ & $y_8$ & $z_0$ \\
   & 100x & $g_5 \Rightarrow g_{15} \Rightarrow \cdots \Rightarrow z_0$ & $y_9$ & $z_0$ \\
$q_4$  & 00xx & $g_{10} (y_0) \Rightarrow \cdots \Rightarrow z_0$ & $y_0$ & $z_0$ \\
   & 01xx & $g_{13} (y_2) \Rightarrow \cdots \Rightarrow z_0$ & $y_2$ & $z_0$ \\
   & 10xx & $g_{11} (y_1) \Rightarrow \cdots \Rightarrow z_0$ & $y_1$ & $z_0$ \\
   & 11xx & $g_{27} (y_3=z_1)$ & $y_3$ & $z_1$ \\
$q_5$  & 00xx & $g_{12} \Rightarrow \cdots \Rightarrow z_0 $ & $y_8$ & $z_0$ \\
   & 10xx & $g_{15} \Rightarrow \cdots \Rightarrow z_0 $ & $y_9$ & $z_0$ \\
$q_6$  & xx00 & $g_{27} (z_1)$ & $y_3$ & $z_1$ \\
   & xx10 & $g_{14} (y_7) \Rightarrow \cdots \Rightarrow z_0 $ & $y_7$ & $z_0$ \\
$q_7$  & xx00 & $g_{10} \Rightarrow \cdots \Rightarrow z_0$ & $y_0$ & $z_0$ \\
   & xx01 & $g_{12} \Rightarrow \cdots \Rightarrow z_0$ & $y_8$ & $z_0$ \\
   & xx10 & $g_9 \Rightarrow \cdots \Rightarrow z_0$ & $y_4$ & $z_0$ \\
$q_8$  & xx00 & $g_{13} \Rightarrow \cdots \Rightarrow z_0$ & $y_2$ & $z_0$ \\
   & xx10 & $g_{16} \Rightarrow \cdots \Rightarrow z_0$ & $y_6$ & $z_0$ \\
\hline
\end{tabular}
\end{scriptsize}
\end{table}

\subsubsection{Analysis}
At first sight of Fig. \ref{figHotBrain} a rough division between
output function generator and checking logic is discernible.
However the ``checker'' reads the output of internal $g_3$ (used to calculate POs $y_8$ and $y_9$) and primary output $y_3$ is reused as error rail $z_1$: hence \emph{the design does not adopt the function-checker
structural constraint}.  

The NXOR tree fed by most primary outputs is akin to parity-checking.  Early attempts at parity grouping \cite{de:94} synthesised each output independently to ensure no sharing within a parity group.  In \cite{toub:97} logic sharing is allowed between outputs of different parity groups.  Circuit A is a highly sophisticated design which goes a step further allowing logic sharing between the parity group (its parity computed at $g_{25}$) and a kind of parity predict (at $g_{24}$) such that for any input word, errors will only be propagated to either the group or the predict.  In fact the whole operation of circuit A is best understood by looking at path sensitisation: for each input word a path is sensitised from each PO line to one and only one of $z_0$ or $z_1$ (sometimes through other outputs) resulting in a fault-secure circuit (Table \ref{hotbrainFS}).  Just as it is challenging to understand biological organisms designed by a similar stochastic evolutionary process; so can it be to apply our rational `divide \& conquer' intellect to understanding an evolved `holistic' circuit. A full understanding and account of its operation is by all means possible but would exceed the limits of this publication.  However we will attempt to cover the core operation as follows:

The difference between the scheme adopted and standard parity-checking is that an explicit parity bit is not generated to always ensure even or odd parity, instead a parity bit $g_{24}$ (calculated using $y_5$) is added to the group $y_{0-2},y_{4},y_{6-9}$ such that their overall parity will always be opposite to $y_3=z_1$ (see Table \ref{hotbrainfaultfree}).  Efficiency is gained by using another output ($y_5$) to generate the parity bit and in its result being used directly as a fluctuating error rail opposite to another existing output $y_3=z_1$.  This demonstrates \emph{the synthesis method presented is capable of finding highly efficient ad hoc TSC strategies well suited to the particularities of the circuit function}.  Mathematically $g_{24}$ is a parity predict for $G_y=y_{0-4},y_{6-9}$, however the implementation described allows for direct availability of a changing dual rail error signal.  Other differences with strict parity-checking is the logic sharing described in the previous paragraph (with $y_5$ used both to calculate $y_{8-9}$ and the `predict' $g_{24}$) and the fact that some outputs ($y_3,y_5$) are not directly part of parity groups but are  used for generating check bits.  Observation of an ancestor of circuit A
revealed that it evolved from circuits in which all outputs fed
the NXOR tree directly, showing how evolutionary search is capable of finding solutions in $S_C$ and $S \cap \overline{S_C}$ defined in \S\ref{intro}.  However, parsimony pressure must have
selected variants in which less outputs were directly checked.  The strategy of cascading outputs whilst ensuring adequate error propagation for achieving low overhead self-checking is
present in several evolved designs \citep{garv:03b} such as circuit B.  This strategy is combined with one similar to
parity-checking to achieve efficient CED \emph{demonstrating the
use of multiple strategies within a single circuit}.  Given
the fact that the original design contained many outputs with
paths usually sensitised to only one output at a time,
parity-checking is a well-adapted strategy to the particularities
of this circuit.  The output cascading used is also well-adapted
to the last three primary outputs.  This shows how \emph{the
evolutionary synthesis process is capable of applying multiple
suitable TSC strategies} to the design of a circuit with TSC CED
not adopting the function-checker structural constraint.  

Even though evolved circuits are, as demonstrated, not constrained by modular decomposition, it is sometimes possible to identify structures and extract design principles from them.  For instance 
 $g_1,g_6,g_7,g_8,g_{10-11},g_{13},g_{27}$ can be seen as an `error propagating block' for which a path is ever sensitised from $g_4$ to a single block output.  Such structure aids the flow control of error signals whilst performing useful computation.  As well as being further evidence of well adapted synthesised TSC strategies, such extracted structures could inform or be used as a building blocks in this or other self-checking synthesis processes. 
\section{Conclusions}
\label{conclusions}
\label{digSCConc}
We present the first automatic TSC circuit synthesis method and demonstrate its effectiveness on a range of benchmark circuits. Previous methods do not automatically resolve the problem of their checkers not being ST for circuits with limited output codespaces.  Our method guarantees TSC circuits everytime and requires on average 23\% duplication overhead for MCNC'91 benchmarks of up to 741 literals (the previous best in literature was 69\% for the benchmarks tackled).  Resulting circuits cover all gate input and output faults and do not impose output encoding changes thus having no hidden external interfacing overhead.  The method shows that highly efficient TSC circuits with no modular decomposition between function generating logic and checker exist, and is the first to synthesise them whilst adopting a tailored checking strategy for each circuit.  Due to the stochastic nature of the process, solutions are also `design diverse' which could be of use if multiple versions without common mode failures were required.  The synthesis method is also shown to be able to make existing irredundant circuits TSC  non-intrusively, with on averge 32\% of duplication overhead for the same set of benchmarks, thus opening its potential for IP-cores.  Given the TSC property is thoroughly checked for the overall design, internal module interfacing issues (see \S\ref{intro}) are avoided.

The method is applicable to any technology library (FPGA, CPLD, transistor level, a specific gate set such as G10-p, nano scale) and fault model (such as transition delay faults) and given its capacity to tailor solutions would likely excel no matter what technology was chosen.  The method could also be applied to synthesis of partially self-checking circuits such as \cite{goes:00,moro:00}, and could also include routing area, performance, power consumption \cite{ghos:05} and minimal ST input set fitness metrics.  Should no definite ordering exist between such metrics multiple Pareto optimal solutions could be generated. Previous work
\citep{de:94} has suggested certain self-checking approaches may
be more favourable than others for certain circuits when routing
area is considered even if resulting in a higher gate count.  By including routing area in our fitness metric, $f_p$, our synthesis process would be able to find suitable ad hoc self-checking strategies adapted to the particularities of the circuit at hand and the particular routing constraints of the given technology (eg. FPGA, ASIC).

Future work will tackle larger combinational benchmarks, sequential circuits, and could include investigating whether circuits exist that meet the TSCG without the `all inputs between faults' assumption, and whether this method can synthesise them.  It could also easily be extended to evaluate strong fault secureness \cite{smit:78}, potentially finding smaller circuits meeting the TSCG.  The method could also be seeded with an efficient conventional strategy such as parity grouping.  

Larger circuits will be tackled by optimising the fitness evaluation metrics using \#SAT \cite{dali:13,vasi:14} or a variety of the TESTDETECT \cite{bour:69} algorithm.  Indeed the main limitation of the evolutionary approach to hardware design is the exponential increase in computational effort with problem size.   However, the landscape is rapidly changing in that area. Just as machine learning algorithms which had remained largely unchanged for 30 years are now successfully tackling real-world problems which were previously unfeasible, because of the application of huge grid computing resources, a similar approach could be used in applying our evolutionary methods to larger industrial scale designs. Increased computing resources would allow greater parallelisation of population evaluations as well as a larger distributed grid of islands which would increase the efficiency of the approach.  The limit of 150 processors used in the current work is dwarfed by the processing power now employed in some industrial machine learning applications; it is only a matter of time before very large grid resources will be routinely available, making our approach -- which we have argued is in the spirit of TSC pioneers' original vision -- feasible for much larger circuits.

The method presented could also be used to mitigate the limitation of other techniques by synthesising TSC checkers which are guaranteed ST under limited output codespaces.  Such checkers are well within reach of those the sizes tackled successfuly by our method in this publication.  Thus TSC solutions for circuits of tens of thousands of gates could be achieved by combining our method with one such as \cite{dali:13}.

The heuristic tailoring that state of the art techniques are capable of \cite{toub:97,ghos:05,dali:13} is limited to adjusting parameters within a self-checking approach to suit a particular circuit function.  A GA can tailor the entire self-checking approach.  Most state of the art methods encode outputs and then add a TSC checker to validate codewords.  Evolved designs do away with this cumbersome extra and check the circuit directly in a similar way that TSC checkers and early hand-crafted designs did.  By developing a checking strategy intimate to the primary function logic, a smaller amount of purely checking overhead is required.  In the case where PO generating logic can be modified it is done so in a way that it will be more amenable to checking with a resulting even lower amount of purely checking overhead.

\appendices
\section{Output function generator synthesis using Sis}
\label{appOpt}\label{appSis}
Sis \cite{sent:92} was used to synthesise and optimise the benchmarks from the MCNC'91 test suite into two input gate technology using all possible gates.  The following recommended Xilinx scripts for LUT technology of general acceptance were used (; represents a new line): {\footnotesize \texttt{full\_simplify;  print\_stats -f;  source script.rugged}}.  These three commands were used in order and the last two were repeated until the number of factored literals went up.  The circuit used was that resulting previously to the last application of source script.rugged.  The same procedure was effected without the  {\footnotesize{ \texttt{full\_simplify}}} at the start and the circuit with lowest factored literals was used for the second stage.

The second stage applied the following script: {\footnotesize \texttt{sweep; simplify; sweep; simplify; xl\_split -n 2; sweep; simplify;
xl\_split -n 2; sweep; xl\_partition -n 2; sweep; simplify; xl\_partition -n 2;
sweep; xl\_k\_decomp -n 2; sweep; xl\_cover -n 2 -h 3 }}.  The resulting circuit was then used as the output function generator to seed evolution with and is the one used to calculate duplication overhead.

\bibliographystyle{plain}
\bibliography{refs}

\begin{IEEEbiography}[{\includegraphics[width=1in,height=1.25in,clip,keepaspectratio]{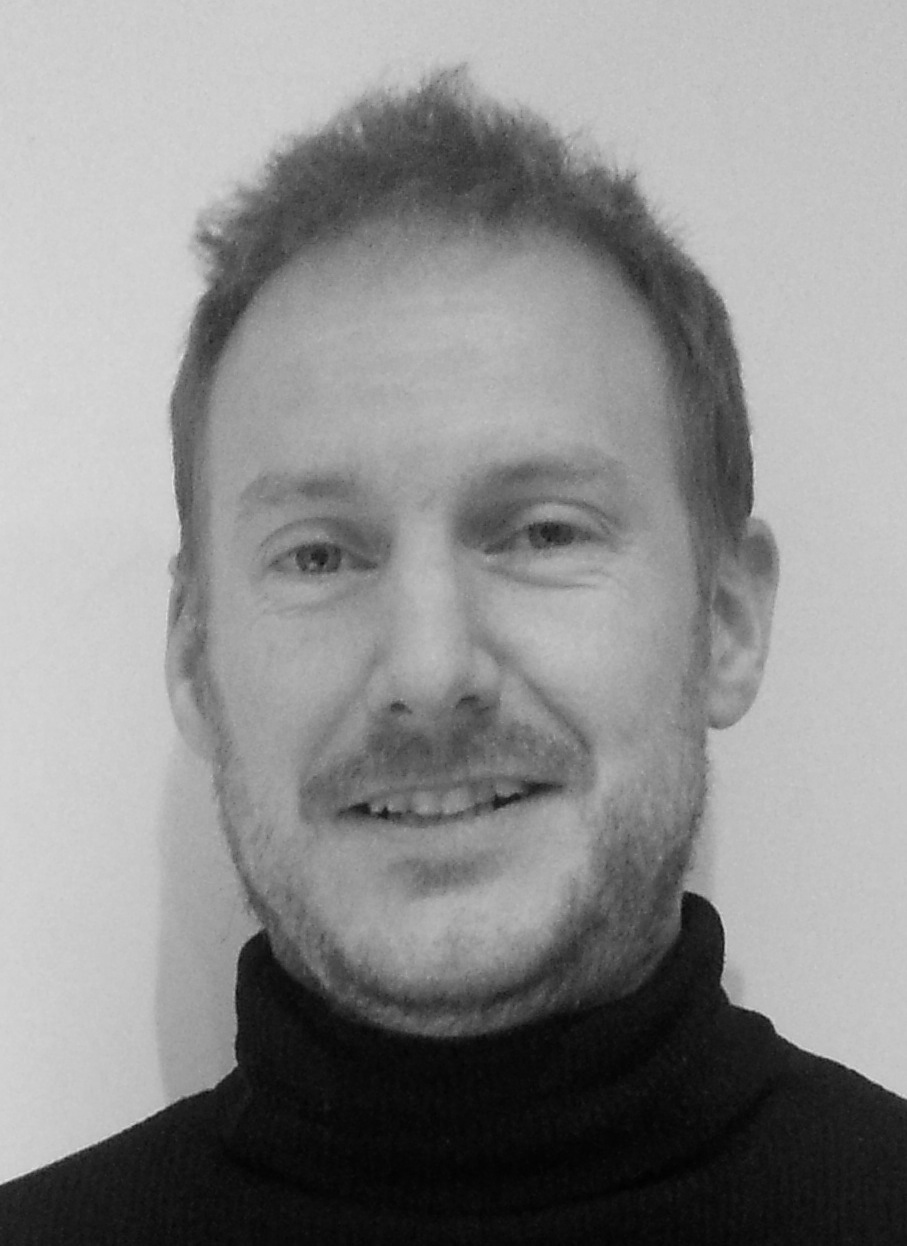}}]{Michael Garvie}
 read Computer Science at the University of Cambridge graduating in 2001 and completing a doctorate degree at the University of Sussex in 2005.  
 
 His research explores the use of evolutionary algorithms for design of analog and digital hardware applied in the fields of concurrent error detection and robotics.
\end{IEEEbiography}
\vskip 0pt plus -1fil
\begin{IEEEbiography}[{\includegraphics[width=1in,height=1.25in,clip,keepaspectratio]{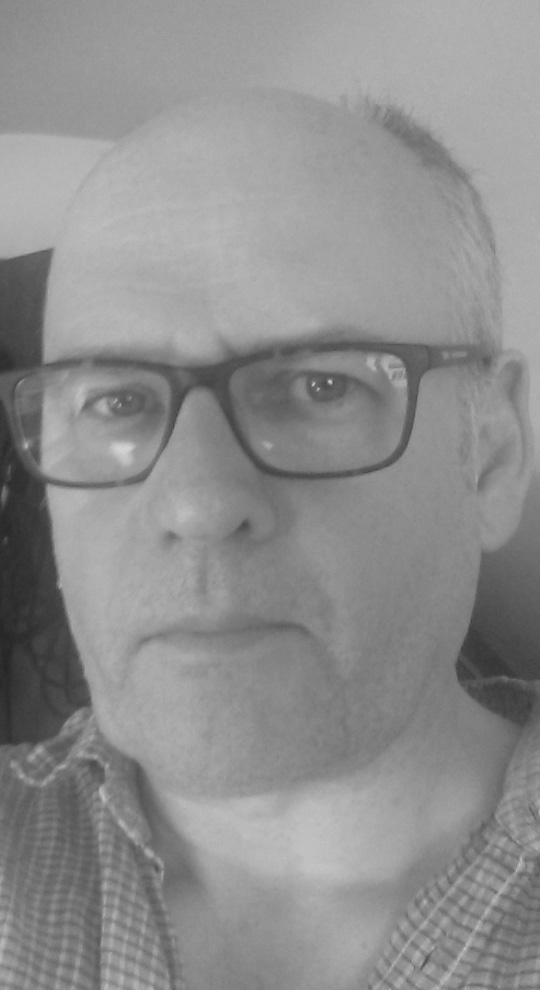}}]
{Phil Husbands} is Research Professor of AI at the University of Sussex and is founding co-director of the Centre for Computational Neuroscience and Robotics. He has a BSc in Physics from Manchester University, a MSc in Computer Engineering from SouthBank University and a PhD in AI from Edinburgh University. 

His research interests include evolutionary systems, applications of evolutionary computing in engineering, biorobotics and computational neuroscience.
\end{IEEEbiography}
\end{document}